\documentclass{ieeeaccess}
\usepackage{cite}
\usepackage{amsmath,amssymb,amsfonts}
\usepackage{algorithmic}
\usepackage{graphicx}
\usepackage{textcomp}

\usepackage{amsmath,graphicx,multirow,booktabs,url,color}

\usepackage{subcaption}

\usepackage[bookmarks=false,colorlinks]{hyperref}

\usepackage{color}

\usepackage{enumitem}

\newcommand{\comment}[1]{}

\def\x{\mathbf{x}}

\def\w{\mathbf{w}}

\def\fpr{\texttt{FPR}}
\def\tpr{\texttt{TPR}}
\def\fp{\texttt{FP}}
\def\tp{\texttt{TP}}
\def\fn{\texttt{FN}}
\def\tn{\texttt{TN}}
\def\lauc{\textit{AUC}}
\def\auc{\texttt{AUC}}

\usepackage{enumitem}

\def\BibTeX{{\rm B\kern-.05em{\sc i\kern-.025em b}\kern-.08em
    T\kern-.1667em\lower.7ex\hbox{E}\kern-.125emX}}
\begin{document}
\history{Date of publication xxxx 00, 0000, date of current version xxxx 00, 0000.}
\doi{10.1109/ACCESS.2017.DOI}

\title{Robust Attentive Deep Neural Network for Detecting GAN-Generated Faces}

\author{
\uppercase{Hui Guo \authorrefmark{1}}, 
\uppercase{Shu Hu \authorrefmark{2}}, 
\uppercase{Xin Wang \authorrefmark{3}} \IEEEmembership{Senior Member, IEEE},
\uppercase{Ming-Ching Chang \authorrefmark{1}} \IEEEmembership{Senior Member, IEEE},
\uppercase{Siwei Lyu \authorrefmark{2}} \IEEEmembership{Fellow, IEEE}
}
\address[1]{Department of Computer Science, College of Engineering and Applied Sciences, University at Albany, State University of New York, Albnay, NY, USA (e-mail: \url{hguo@albany.edu}, \url{mchang2@albany.edu})}
\address[2]{Department of Computer Science and Engineering,
      University at Buffalo, State University of New York,
      Buffalo, NY 14260-2500, USA (e-mail: \url{shuhu@buffalo.edu}, \url{siweilyu@buffalo.edu})}
\address[3]{Keya Medical, Seattle, WA, USA (e-mail: \url{xinw@keyamedna.com})}

\tfootnote{{\bf Acknowledgement:} This work is partly supported by the US Defense Advanced Research Projects Agency (DARPA) Semantic Forensic (SemaFor) program under Kitware, Inc. project {\it Semantic Information Defender (SID)} for State University of New York and Albany and Buffalo with grant K003088-00-S04.}

\markboth
{Guo \headeretal: Robust Attentive Deep Neural Network for Detecting GAN-Generated Faces}
{Guo \headeretal: Robust Attentive Deep Neural Network for Detecting GAN-Generated Faces}

\corresp{Corresponding author: Ming-Ching Chang  (e-mail: \url{mchang2@albany.edu}).}



\begin{abstract}
GAN-based techniques can generate and synthesize realistic faces that cause profound social concerns and security problems. Existing methods for detecting GAN-generated faces can perform well on limited public datasets. However, images from existing datasets do not represent real-world scenarios well enough in terms of view variations and data distributions, where real faces largely outnumber synthetic ones. The state-of-the-art methods do not generalize well in real-world problems and lack the interpretability of detection results. Performance of existing GAN-face detection models degrades accordingly when facing data imbalance issues. To address these shortcomings, we propose a robust, attentive, end-to-end framework that spots GAN-generated faces by analyzing eye inconsistencies. Our model automatically learns to identify inconsistent eye components by localizing and comparing artifacts between eyes. After the iris regions are extracted by Mask-RCNN, we design a {\it Residual Attention Network (RAN)} to examine the consistency between the {\it corneal specular highlights} of the two eyes. Our method can effectively learn from imbalanced data using a joint loss function combining the traditional cross-entropy loss with a relaxation of the ROC-AUC loss via {\it WMW statistics}. Comprehensive evaluations on a newly created {\bf FFHQ-GAN} dataset in both balanced and imbalanced scenarios demonstrate the superiority of our method.
\end{abstract}

\begin{keywords}
GAN-generated face, fake face detection, iris detection, corneal specular highlights, Residual Attention Network, data imbalance, AUC maximization, WMW statistics, FFHQ-GAN dataset.
\end{keywords}

\titlepgskip=-15pt

\maketitle

\section{Introduction}
\label{sec:introduction}

\begin{figure}[t]
\centering
\includegraphics[width=76mm]{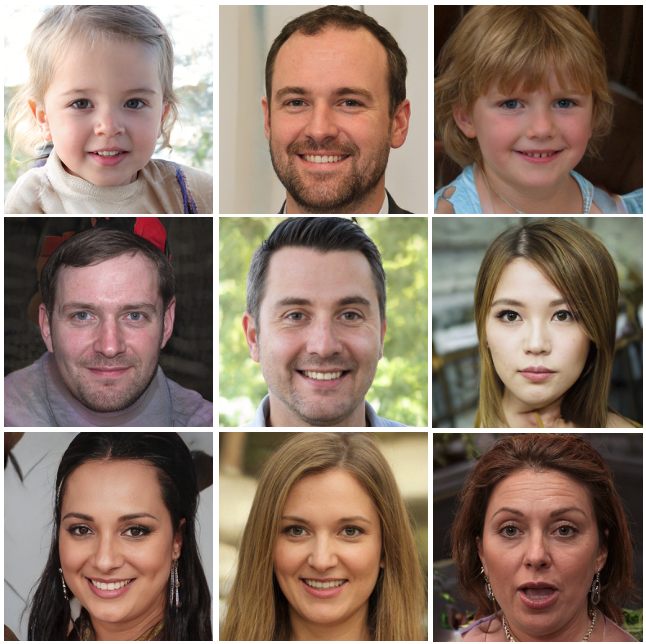}
\caption{StyleGAN2~\cite{karras2020analyzing} generated faces are highly realistic and can be easily abused for malicious purposes. Effective forensic methods for identifying them is of strong needs.
}
\label{fig:introfig}
\vspace{-1em}
\end{figure}

The development of Generative Adversarial Networks (GANs)~\cite{Goodfellow2014Generative} has led to a dramatic increase in the realism in generating high-quality face images, including PGGAN~\cite{karras2017progressive}, StyleGAN~\cite{karras2019style}, StyleGAN2~\cite{karras2020analyzing}, and StyleGAN3~\cite{Karras2021}. As illustrated in Figure~\ref{fig:introfig}, these GAN generated (or synthesized)  fake faces are difficult to distinguish from human eyes. Such synthesized faces are easily generatable, can be directly leveraged for disinformation, and potentially lead to profound social, security, and ethical concerns. 
The GAN-generated faces can be easily abused for malicious purposes, such as creating fake social media accounts to lure or deceive unaware users~\cite{theverge,cnn1,cnn2,reuters}, which can cause significant security problems and frauds. Therefore, the authentication of GAN-generated faces has obtained increasing importance in recent years. However, there exists only a paucity of forensic techniques that can effectively detect such fake faces.

Many studies employ CNNs or other classifiers to distinguish the GAN-generated faces from the real ones~\cite{yang2019exposing,marra2019gans,mo2018fake,do2018forensics,wang2019fakespotter,chen2021locally}. Although these methods detect various GAN-generated faces with relatively high accuracy, similar to other deep learning-based techniques, they suffer from poor generalization and lack interpretability of detection results. Physiology-based methods~\cite{matern2019exploiting,yang2019exposing,guo2021eyes} detect fake faces by examining the semantic aspects of human faces, including physiological or shape-related cues such as symmetry, iris color, and pupil shapes.
Our prior work of an explainable physical method in \cite{hu2021exposing} addressed some of the above limitations, where GAN-generated faces are identified based on a rule-based decision over the inconsistencies of the specular eye patterns. However, this method relies on assumptions of a frontal face as input and the existence of far-away lighting reflection source(s) from both eyes. When these assumptions are violated, generalization will be limited and false positives may rise significantly.


In this paper, we improve our prior method of \cite{hu2021exposing} and develop an end-to-end approach for detecting GAN-generated faces by examining the inconsistencies between the corneal specular highlights of the two eyes. We first use Mask R-CNN \cite{he2017mask} to detect and localize the iris regions. Instead of segmenting the corneal specular highlights using low-level image processing methods in \cite{hu2021exposing}, we design a {\bf Residual Attention Network} which consists of {\it residual attention blocks} inspired from \cite{wang2017residual}, to automatically learn to localize the inconsistencies. Our new method is data-driven and can better spot inconsistent artifacts, including but not limited to the corneal specular highlights.

{\bf Data imbalance} is an important issue that is less addressed in existing GAN-generated face detection works. In real-world use scenarios of face examination, real face images usually outnumber GAN-generated ones by a large amount. Imbalanced data lead to learning problems and thus affect model design. It is well-known the widely-used cross-entropy loss~\cite{murphy2012machine} is not suitable for classifying imbalanced data. Although substantial progress is made by sampling~\cite{szegedy2015going}, adjusting of class weights, data enhancement~\cite{fadaee2017data}, {\it etc.}, learning with imbalanced data is still challenging. It is intuitive that the Area Under Curve (AUC) of the Receiver Operating Characteristic (ROC) plot can be incorporated as a loss term to improve classification performance~\cite{provost1998case}. However, the AUC is a pairwise rank-based metric with discontinuous values among iterations. Therefore, AUC is not directly applicable for loss design for end-to-end optimization of the classifier.
To this end, we incorporate a ROC-AUC loss term by maximizing the Wilcoxon-Mann-Whitney (WMW) statistics of the ROC, which is shown to provide similar effects in approximating the AUC optimization~\cite{yan2003optimizing,lyu2018univariate}. Our experimental results show that a combination of the binary cross-entropy loss and the WMW-AUC loss leads to the best end-to-end result.


We perform experiments on two data sources: (1) 
real human face images obtained from the Flickr-Faces-HQ (FFHQ) dataset \cite{karras2019style} and 
(2) GAN-generated face images available from \url{http://thispersondoesnotexist.com} as shown in Figure~\ref{fig:introfig}. Experiment results demonstrate the superiority of the proposed method in distinguishing GAN-generated faces from the real ones.
We summarize the main contributions of this paper in the following:
\begin{itemize}[leftmargin=16pt] \itemsep 0.5em

\item We propose an end-to-end method for detecting GAN-generated faces by visually comparing the two eyes. A residual attention network model is incorporated to better focus on the inconsistencies of the eyes {\it e.g.} corneal specular highlights and other artifacts. Our fake face detection method is interpretable, and the proposed cues can be leveraged by human beings as well to perform examinations.

\item We introduce the WMW-AUC loss that approximates the direct optimization of the AUC. This can also effectively address the data imbalance learning problem in contrast to other sampling or data augmentation approaches.

\item We generate a new {\bf FFHQ-GAN} dataset by combining portions of the FFHQ real faces with the StyleGAN2 generated images. Performance of GAN-generated face detection is evaluated on this FFHQ-GAN dataset for both balanced and imbalanced data conditions. Experimental results show that our method achieves plausible performance especially on imbalanced datasets. Ablation study also validates the effectiveness of the proposed attention module and the loss design.

\end{itemize}

The paper is organized as follows. Section~\ref{sec:related:work} summarizes related works on GAN-generated faces detection, attention methods, and learning from imbalanced data. Section~\ref{sec:method} describes the proposed network architecture and proposed loss terms for robust learning. Section~\ref{sec:exp} shows experimental results with qualitative visualization and quantitative analysis. Section~\ref{sec:con} concludes this work.

\begin{figure*}[t]
\centering
\includegraphics[width=\linewidth]{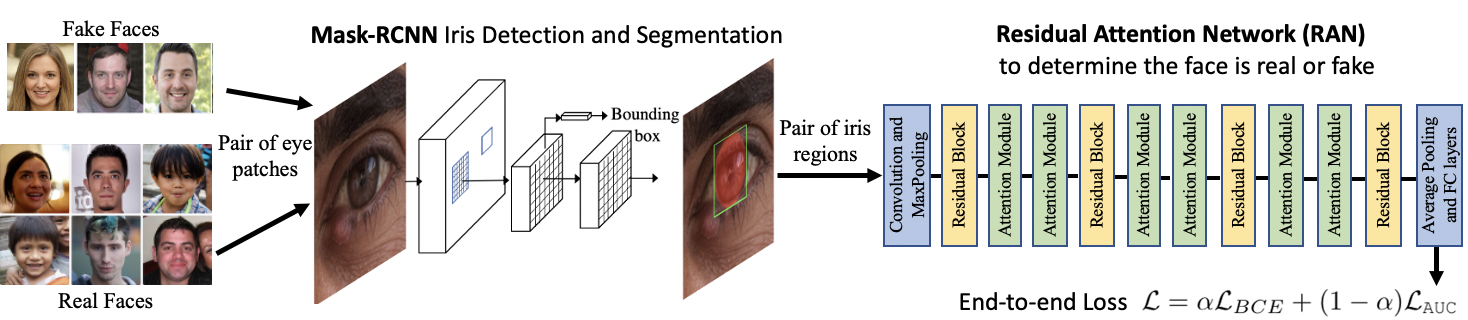}
\caption{{\bf The proposed architecture for GAN-generated face detection.} We first use DLib~\cite{dlib09} to detect faces and localize eyes, and use Mask R-CNN~\cite{he2017mask} to segment out the iris regions. A {\it Residual Attention Network (RAN)} then performs binary classification on the extracted {\it iris pair} to determine if the face is real or fake. The training is carried out using a joint loss combining the {\it Binary Cross-Entropy (BCE) loss} and the {\it ROC-AUC loss with WMW relaxation} to better handle the learning from imbalanced data (see text). 
}
\label{fig:network}
\vspace{-0.3cm}
\end{figure*}

\section{Related Works}
\label{sec:related:work}

We briefly review related works, including GAN-generated face detection methods. We also review the literature on the attention mechanism and learning from imbalanced data.

\subsection{GAN-generated face detection}

GAN-generated faces detection methods can be organized into two categories. 

{\bf Data-driven methods}~\cite{marra2019incremental,goebel2020detection,wang2020cnn,liu2020global,hulzebosch2020detecting} mostly train a deep neural network model to distinguish real and GAN-generated faces. These deep learning (DL) based methods work well in many scenarios, as they can better learn representations in a high-dimensional feature space instead of raw image pixels.

{\bf Physical and physiological methods} look for signal traces, artifacts, or inconsistencies left by the GAN synthesizers. These methods are explainable in nature. 
Simple cues such as color difference are used in \cite{mccloskey2018detecting,li2018detection} to distinguish GAN images from the real ones. However, those methods are no longer effective as the GAN methods advance. 
More sophisticated methods in \cite{marra2019gans, yu2019attributing} leverage fingerprints or abstract signal-level traces of the noise residuals to differentiate GAN-generated faces.
Many works \cite{zhang2019detecting,frank2020leveraging,durall2020watch} identify GAN images by recognizing the specific artifacts produced by the GAN upsampling process.
In \cite{yang2019exposinggan}, the distribution of facial landmarks is analyzed to distinguish GAN-generated faces. Inconsistent head poses are detected to expose the fake videos in \cite{yang2018exposing}. The work of \cite{matern2019exploiting} identifies GAN-generated faces as well as deepfakes face manipulations by inspecting visual artifacts. 
Our prior work of \cite{hu2021exposing} determines the inconsistencies of the corneal specular highlights between left and right eyes to expose GAN-generated faces.

\subsection{Attention Mechanism}

Since the seminal work of \cite{bahdanau2014neural} in machine translation, the {\em attention mechanism} is widely used in many applications on improving the performance of deep learning models by focusing on the most relevant part of the features in a flexible manner. The Class Activation Mapping (CAM)~\cite{zhou2016learning} and Grad-CAM~\cite{selvaraju2017grad} are widely used in many computer vision tasks~\cite{chattopadhay2018grad}. However, in these works, attentions are only used to visualize model prediction in showing significant portions of the images. On the other hand, integrating the attention mechanism into the network design is shown to be effective in boosting performance, as the network can be guided by the attention to focus on relevant regions during training~\cite{kardakis2021examining}. 

The channel attention~\cite{CBAM:ECCV2018} can automatically learn to focus on important channels by analyzing the relationship between channels. SENet~\cite{hu2018squeeze} embeds the channel attention mechanism into residual blocks, and effectiveness is shown on large-scale image classification. The attention mechanism is also used in \cite{zhang2018image} to distinguish important channels in the network to improve the representation capability.
The idea of channel attention and spatial attention are combined jointly in \cite{chen2017sca,woo2018cbam} to improve network performance significantly. The Residual Attention Network in \cite{wang2017residual} combines the residual unit \cite{he2016deep} with the attention mechanism by stacking residual attention blocks to improve performance and reduce model complexity.


\subsection{Imbalanced Data Learning}

Learning from imbalanced data has been widely studied in machine learning~\cite{yang2020rethinking, cao2019learning, hu2020learning, hu2021sum} and computer vision~\cite{wang2019dynamic, huang2016learning}. Earlier solutions for imbalanced data learning are mainly based on the sampling design, {\em e.g. } {\em oversampling} for minor classes, {\em undersampling} for major classes, and weighed sampling~\cite{he2009learning}, {\em etc.} These sampling-based methods come with their own drawbacks. For example, undersampling may ignore important samples, and oversampling may lead to overfitting. 

Data augmentation provides an alternative solution to alleviate data imbalance issues. For image recognition, image mirroring, rotation, color adjustments, {\em etc.} are simple methods to augment data samples \cite{shorten2019survey}. However, data augmentation methods can only address the data imbalance problems partly, as the size of the original dataset must be diverse enough, such that a sufficient amount of representative samples can be produced from augmentation.

\section{Method}
\label{sec:method}

\begin{figure*}[t]
\centering
\includegraphics[width=0.9\linewidth]{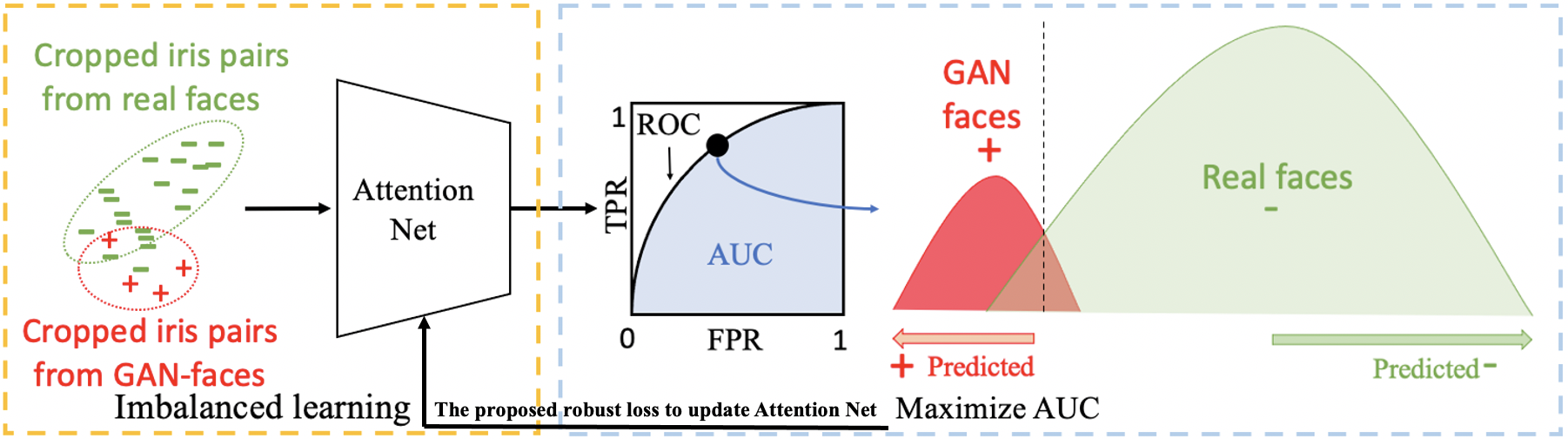}
\caption{{\bf The proposed pipeline for training the Residual Attention Network (RAN) on possibly imbalanced data for GAN-generated face classification.} The extracted iris pairs are passed as input to the RAN. A robust loss function derived from maximizing the AUC of ROC is optimized in the training of the RAN. See details in $\S$~\ref{sec:WMW:AUC}.}
\label{fig:framework}
\vspace{-0.3cm}
\end{figure*}

We next describe the proposed GAN-generated faces detection framework.
Given an input face image, facial landmarks are first localized using DLib~\cite{dlib09} and Mask R-CNN~\cite{he2017mask} is used to segment out the left and right iris regions of the eyes ($\S$~\ref{sec:iris:det}).
We adopt a residual attention based network \cite{wang2017residual} to perform binary classification on the iris regions of interest to determine if the input image is real or fake ($\S$~\ref{sec:residual:attn:net}). 
The training of our network aims to maximize the classification performance reflected by the standard Area-Under-Curve of the ROC plot ($\S$~\ref{sec:roc:auc}), which is general and can effectively address the data imbalance problem.
However, due to the discrete nature of the ROC-AUC values, a naive gradient-based implementation does not work for end-to-end learning. 
In $\S$~\ref{sec:WMW:AUC}), we present a detailed solution in our proposed approach by relaxing the AUC maximization and approximating the goal using the WMW statistics.
Figure~\ref{fig:framework} overviews the training pipeline of the proposed method that can effectively learn from imbalanced data.

\begin{figure}[t]
\centering
\includegraphics[width=85mm]{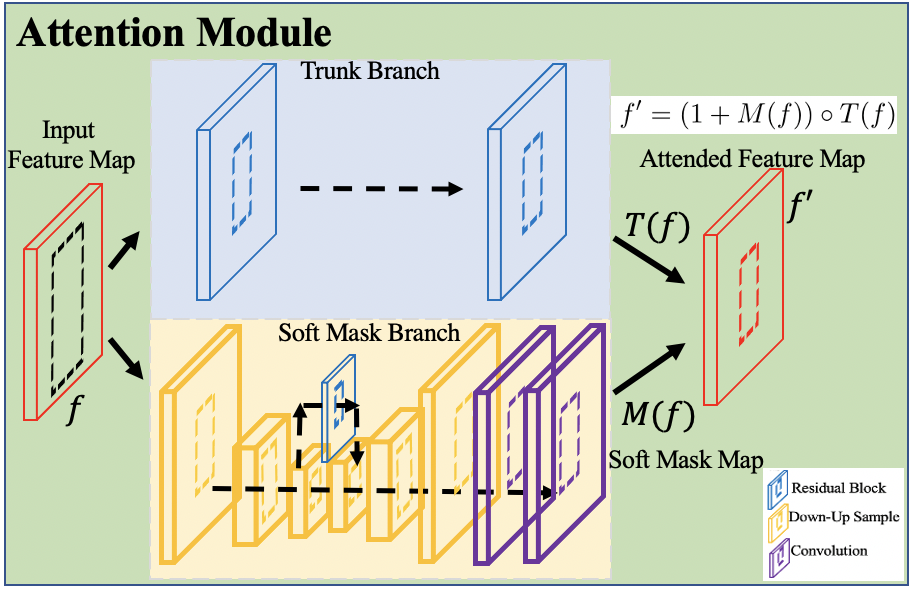}
\caption{{\bf Details of our Attention Module} from the our RAN in Figure~\ref{fig:network}. Our design is inspired from the residual attention network of \cite{wang2017residual}.
}
\label{fig:attnet}
\vspace{-0.3cm}
\end{figure}

\subsection{Facial Landmark Localization and Iris Segmentation}
\label{sec:iris:det}

\begin{figure}[t]
\centerline{
\includegraphics[height=0.4\linewidth]{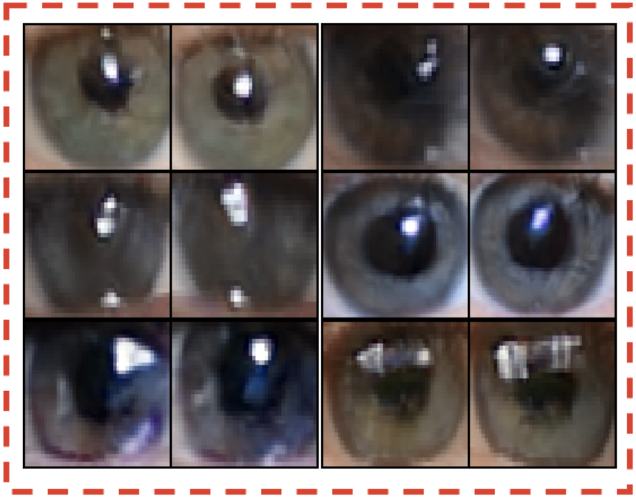}
\includegraphics[height=0.4\linewidth]{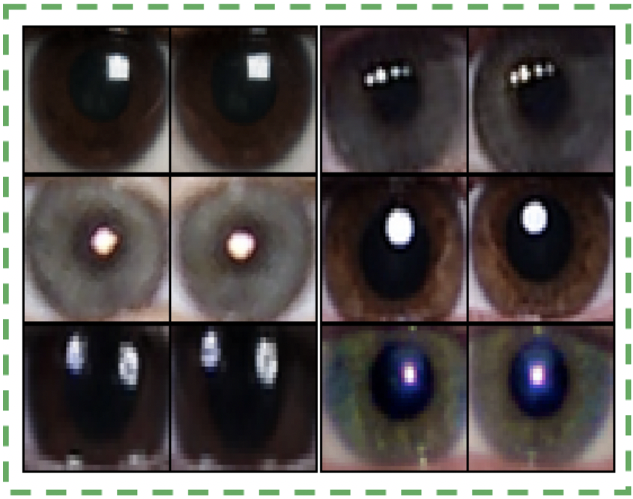}
\vspace{-0.1cm}
}
\centerline{
{\footnotesize \hspace{-0.6cm} (a) from {\bf GAN-generated} faces  \hspace{1.5cm}
(b) from {\bf real} faces.}
}
\caption{{\bf The extracted iris pairs} of our method for the (a) GAN-generated and (b) real faces. 
Artifacts of inconsistent corneal specular highlights are obvious in GAN-generated iris pairs.}
\label{fig:iris}
\vspace{-1em}
\end{figure}

Given a face image, the first step our method to determine if it is real or GAN-generated is to detect and localize the face using the facial landmark extractor provided in DLib~\cite{dlib09}. The localized regions containing the eyes are cropped out for consistency checking.
Mask R-CNN~\cite{he2017mask}, the state-of-the-art detection and segmentation network, is employed to further detect and localize the iris regions.
Mask R-CNN is a two-stage network based on Faster R-CNN~\cite{ren2015faster} as shown in Figure~\ref{fig:network} (middle). 
The first stage of Mask R-CNN is a Region Proposal Network (RPN) that generates candidate object bounding boxes for all the object categories. 
In the second stage, the R-CNN  extracts features using the Region of Interest Align (RoIAlign) layer for each proposal.
In the last stage, label classification and bounding box regression are performed for each proposal, and mask prediction is performed in a parallel branch.
We train the Mask R-CNN model using the eye region images from the datasets in \cite{wang2020iris, wang2021nir}, where more details will be provided in $\S$~\ref{sec:implement}. 
Figure~\ref{fig:iris} shows examples of the extracted {\it pair of iris regions} from the cases of GAN-generated (left) and real (right) faces.


\begin{table}[t]
\caption{Details of the proposed {\bf Residual Attention Network (RAN)} in the right of Figure~\ref{fig:network}. }
\centering 
\begin{tabular}{c|c|c}
\hline
Layer             & Output Size      & Network       \\ \hline
Conv1             & 96$\times$96$\times$16        &{3 $\times$ 3, stride 1} \\ \hline
Max pooling       & 48$\times$48$\times$16        & {3 $\times$ 3, stride 2}      \\ \hline
Residual block     & 48$\times$48$\times$64        & {
$1\times
\begin{bmatrix}
1 \times 1 , 16\\ 
3 \times 3 , 16\\ 
1 \times 1  , 64\\
1 \times 1 , 64
\end{bmatrix}$
}                    \\ \hline
Attention Module  & 48$\times$48$\times$64        & 2$\times$ attention           \\ \hline
Residual block     & 24$\times$24$\times$128       & {
$1\times
\begin{bmatrix}
1 \times 1  , 32\\ 
3 \times 3  , 32\\ 
1 \times 1  , 128\\
1 \times 1  , 128
\end{bmatrix}$
}                    \\ \hline
Attention Module  & 24$\times$24$\times$128        & 2$\times$ attention              \\ \hline
Residual block     & 12$\times$12$\times$256        & {
$1\times
\begin{bmatrix}
1 \times 1 , 64\\ 
3 \times 3 , 64\\ 
1 \times 1 , 256\\
1 \times 1 , 256
\end{bmatrix}$
}                    \\ \hline
Attention Module  & 12$\times$12$\times$256        & 2$\times$ attention             \\ \hline
Residual block     & 6$\times$6$\times$512          & {
$3\times
\begin{bmatrix}
1 \times 1 , 128\\ 
3 \times 3 , 128\\ 
1 \times 1 , 512\\
1 \times 1 , 512
\end{bmatrix}$ 
}                    \\ \hline
Average pooling   &  1$\times$1$\times$512         & {6 $\times$ 6, stride 1}       \\ \hline
FC, Sigmoid        & {1}                                  \\ \hline

\end{tabular}
\label{Table:architectures}
\vspace{-1em}
\end{table}

\subsection{Residual Attention Network}
\label{sec:residual:attn:net}

We adopt the attention mechanism~\cite{chen2016attention,jaderberg2015spatial,jin2020ra,CBAM:ECCV2018,zhang2018image} to improve the spotting of inconsistent corneal specular highlights between the eyes so as to improve GAN-generated face detection. 
Incorporating attention to a detection/segmentation network is commonly accomplished by having a separate branch that calculates the attention maps and later being incorporated back to the main branch with weights.
Inspired from \cite{wang2017residual}, each Attention Module in our attention network consists of a {\it trunk branch} and a {\it soft mask branch}. 
The trunk branch contains several residual blocks~\cite{he2016deep} and acts as a shortcut for data flow. The soft mask branch uses a U-net structure~\cite{ronneberger2015u} to weight output features. 
Specifically, given the input feature map ${f}$, denote the output of the trunk branch ${T}$ as ${T}({f})$, and the output of the soft mask branch ${M}$ as ${M}({f})$, respectively. 
As illustrated in Figure~\ref{fig:attnet}, the final attended feature map ${f}^{\prime}$ is obtained via element-wise matrix product via 
\begin{equation}
{f}^{\prime}= \left( 1 + {M}({f})) \circ {T}({f} \right),
\end{equation}
where the symbol $\circ$ denotes the Hadamard product.


The attention module can be configured to focus learning on channel attention, spatial attention, or mixed attention. As suggested in \cite{wang2017residual}, the mixed attention yields the best performance. Thus, we use the Sigmoid function $\frac{1}{1+\exp \left(-f_{s, c}\right)}$  to learn the mixed attention for each channel and each spatial location,
where $s$ ranges over all spatial positions and $c$ ranges over all channels of $f$. 
The proposed {\bf Residual Attention Network (RAN)} is constructed by stacking multiple Attention Modules, as shown on the right side of Figure~\ref{fig:network}.
Table~\ref{Table:architectures} provides details of the architectures.
Although the attention module plays an important role in classification, a simple stacking of attention modules may reduce performance. 
To this end, we adopt a simple solution by adding the attention map onto the original feature map. This combination allows attention modules stacked like a ResNet \cite{he2016deep} and improves the performance \cite{wang2017residual}.
Given an input image, The RAN outputs a prediction score from the last Sigmoid layer, as an indication of the likelihood of the input image being a GAN-generated image.




\subsection{AUC of ROC for Classification Evaluation}
\label{sec:roc:auc}

Most classification loss measures including the popular cross-entropy loss are ineffective in addressing the issue of data imbalance. The resulting models can produce accurate but rather biased predictions that do not work well in practice. It is desirable to address data imbalance directly by specifically designing a suitable loss function. 

Since the area under the curve (AUC) of a receiver operation curve (ROC) \cite{cortes2003auc,lyu2018univariate} is a robust evaluation metric for both balanced and imbalanced data, we would like to directly maximize the  AUC to handle imbalanced situations.
The AUC is widely used in the binary classification problems. We next briefly review the definition of AUC, and then motivate how we incorporate a loss term that directly maximize the AUC performance. Given a labeled dataset $\{(\x_i, y_i)\}_{i=1}^M$, where each data sample $\x_i\in \mathbb{R}^d$ and each corresponding label $y_i\in\{-1, +1\}$. We define a set of indices of positive instances as $\mathcal{P}=\{i \; | \; y_i = +1\}$. Similarly, the set of indices of negative instances is $\mathcal{N}=\{i \; | \; y_i = -1\}$.
Let $g_{\w}:\mathbb{R}^d\rightarrow \mathbb{R}$ be a parametric prediction function with parameter $\w \in \mathbb{R}^m$. 
$g_{\w}(\x_i)$ represents the prediction score of the $i$-th sample, where $i\in\{1,\cdots, M\}$. For simplicity, we assume $g_{\w}(\x_i)\neq g_{\w}(\x_j)$ for $i\neq j$ (ties can be broken in any consistent way). 

Given a threshold $\lambda$, the number of negative examples with prediction scores larger than $\lambda$ is false positive (\fp), and the number of positive examples with prediction scores greater or equal to $\lambda$ is true positive (\tp). According to the \fp~and \tp, we can calculate the false positive rate (\fpr) and the true positive rate (\tpr) as follows,
\begin{equation*}
    \begin{aligned}
    &\fpr=\frac{\sum_{i\in \mathcal{N}}\mathbb{I}_{[g_{\w}(\x_i)>\lambda]}}{|\mathcal{N}|}, \ \ \ \tpr=\frac{\sum_{i\in \mathcal{P}}\mathbb{I}_{[g_{\w}(\x_i)\geq\lambda]}}{|\mathcal{P}|},
    \end{aligned}
\end{equation*}
where $\mathbb{I}_{[a]}$ is an indicator function with $\mathbb{I}_{[a]}=1$ if $a$ is true and 0 otherwise.
The receiver operation curve (ROC) is a plot of \fpr~versus \tpr~with setting different decision thresholds $\lambda\in(-\infty,\infty)$. Based on this definition, ROC is a curve confined to $[0,1]\times [0,1]$ and connecting the point (0,0) to (1,1). The value of \auc~corresponds to the area enclosed by the ROC curve. 







\subsection{WMW AUC Relaxation for Loss Design}
\label{sec:WMW:AUC}

The computation of an AUC score based on the area under a ROC curve cannot be directly used in a loss function, due to its discrete nature. According to the Wilcoxon-Mann-Whitney (WMW) statistic \cite{yan2003optimizing}, we can relax the AUC as follows,
\begin{equation*}
\auc= \frac{1}{|\mathcal{P}||\mathcal{N}|}\sum_{i\in \mathcal{P}}\sum_{j\in \mathcal{N}}\mathbb{I}_{[g_{\w}(\x_i)>g_{\w}(\x_j)]}.
\end{equation*}
Therefore, the corresponding \auc~loss (risk) can be defined as: 
\begin{equation}
\mathcal{L}_{\lauc}=1-\auc= \frac{1}{|\mathcal{P}||\mathcal{N}|}\sum_{i\in \mathcal{P}}\sum_{j\in \mathcal{N}}\mathbb{I}_{[g_{\w}(\x_i)<g_{\w}(\x_j)]}.
\label{eq:auc_loss}
\end{equation}
Obviously, $\mathcal{L}_{\lauc}$ takes value in $[0, 1]$. It is a fraction of pairs of prediction scores from the positive sample and negative sample that are ranked incorrectly, {\em i.e.}, the prediction score from a negative sample is larger than the prediction score from a positive sample. If all prediction scores from the positive samples are larger than any prediction score from the negative samples, then $\mathcal{L}_{\lauc}=0$. This indicates we obtain a perfect classifier. Furthermore, $\mathcal{L}_{\lauc}$ is independent of the threshold $\lambda$. 
$\mathcal{L}_{\lauc}$ only depends on the prediction scores $g_{\w}(\x)$. In other words, the predictor $g_{\w}$ affects the value of $\mathcal{L}_{\lauc}$. Therefore, we aim to learn a classifier $g_{\w}$ that minimizes Eq.\eqref{eq:auc_loss}. 

Although we can calculate $\mathcal{L}_{\lauc}$ by comparing prediction score from the positive sample and prediction score from the negative sample in each pair, the $\mathcal{L}_{\lauc}$ formulation is non-differentiable due to the discrete computation. It is therefore desirable to find a differentiable approximation for $\mathcal{L}_{\lauc}$. Inspired by the work in \cite{yan2003optimizing}, we find an approximation to $\mathcal{L}_{\lauc}$ that can be directly applied to our objective function to minimize the \auc~loss along with our imbalanced training procedure. Specifically, a differentiable approximation of $\mathcal{L}_{\lauc}$ can be reformulated as:
\begin{equation}
\mathcal{L}_{\lauc}= \frac{1}{|\mathcal{P}||\mathcal{N}|}\sum_{i\in \mathcal{P}}\sum_{j\in \mathcal{N}}R(g_{\w}(\x_i),g_{\w}(\x_j)),
\label{eq:relax_auc}
\end{equation}
and $R(g_{\w}(\x_i),g_{\w}(\x_j)) = $
\begin{equation}\small
\begin{aligned}
&\left\{\begin{matrix}
(-(g_{\w}(\x_i) -g_{\w}(\x_j) - \gamma))^p, & g_{\w}(\x_i) -g_{\w}(\x_j)< \gamma ,\\ 
 0,& \mbox{otherwise},
\end{matrix}\right.
\end{aligned}
\label{eq:gamma}
\end{equation}
where $\gamma \in (0,1]$ and $p > 1$ are two hyperparameters. 


\bigskip
{\bf Loss for the proposed Residual Attention Network.}
We use a joint loss function comprising the conventional binary cross-entropy (BCE) loss function $\mathcal{L}_{BCE}$ and the AUC loss function $\mathcal{L}_{\lauc}$ in weighted sum: 
\begin{equation}
\mathcal{L} =  \alpha \; \mathcal{L}_{BCE}+ (1- \alpha) \; \mathcal{L}_{\lauc},
\label{eq:jointloss}
\end{equation}
\noindent where $\alpha\in[0,1]$ is a scaling factor that is designed for balancing the weights of the BCE loss and the AUC loss. 



\begin{figure*}[t!]
\captionsetup[subfigure]{justification=centering}

\centering
\begin{subfigure}[b]{\textwidth}
                \includegraphics[width=\linewidth]{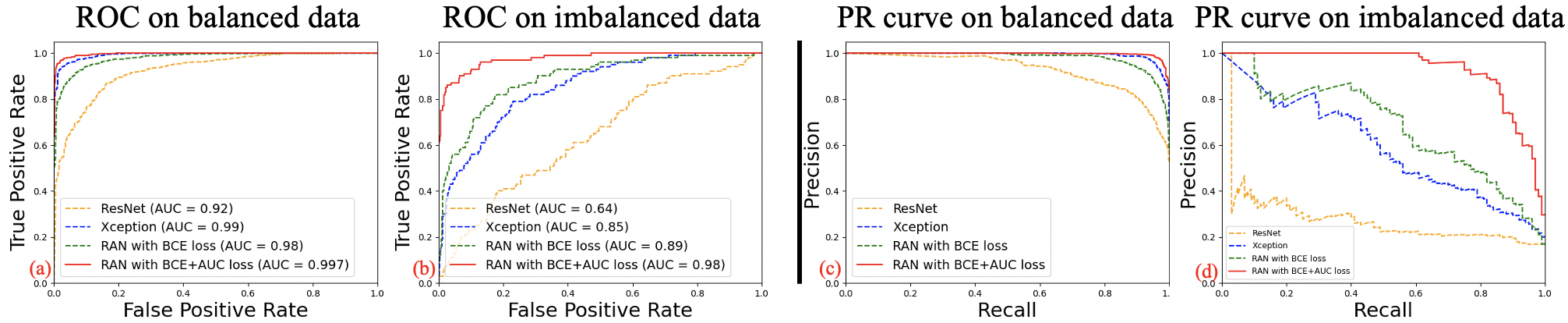}
\end{subfigure}%
\caption{ Performance comparison of the proposed method with ResNet with BCE loss, Xception with BCE loss, and RAN with BCE loss.}

\label{fig_compareRes}
\vspace{-0.5em}
\end{figure*}

\begin{figure*}[t]
\centerline{
  \includegraphics[width=\linewidth]{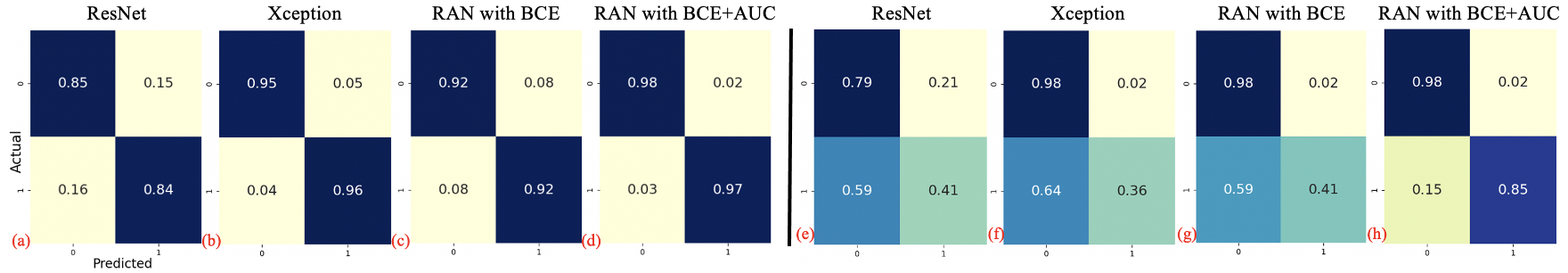}
}  
         
\vspace{-0.5em}
\caption{Confusion Matrix on the FFHQ-GAN (left) balanced and (right) imbalanced datasets.
}
\label{fig_cm_blance}
\vspace{-1em}
\end{figure*}

\section{Experiment}
\label{sec:exp}

For experimental evaluation of the proposed method and comparison against the state-of-the-art methods, we first introduce the newly constructed FFHQ-GAN datasets $\S$~\ref{sec:dataset}. 
Implementation details of the proposed method are provided in $\S$~\ref{sec:implement}. 
Performance evaluation on the FFHQ-GAN balanced and imbalanced subsets is in $\S$~\ref{sec:FFHQ:GAN:result}.
Ablation studies are provided in $\S$~\ref{sec:ablation}.
Finally, qualitative results are shown in $\S$~\ref{sec:qualitative}.


\subsection{The New FFHQ-GAN Dataset} 
\label{sec:dataset}

We collect real human face images from the Flickr-Faces-HQ (FFHQ) dataset \cite{karras2019style}. GAN-generated face images are created using StyleGAN2 \cite{karras2019style} via \url{http://thispersondoesnotexist.com}, where the image resolution is 1024 $\times$ 1024 pixels. 
We randomly select 5,000 real face images from FFHQ and 5,000 GAN-generated face images. After iris detection, we discard those images with the iris of any eye not detected. This ends up with 3,739 real faces (with iris pairs) and 3,748 fake faces (with iris pairs), which constitutes our new {\bf FFHQ-GAN dataset}. The split ratio of training and testing is 8:2.

To enable a thorough evaluation of the model in both balanced and imbalanced data scenarios,  
we sampled the FFHQ-GAN dataset to form an imbalanced subset, where the statistics of the subsets are provided in Table \ref{tab:dataset}.

\begin{table}[t]
\centering
\caption{Details of the FFHQ-GAN dataset regarding its balanced (-b) and imbalanced (-imb) subsets.}
\resizebox{\linewidth}{!}{
\setlength{\tabcolsep}{0.2em} 
\begin{tabular}{|c|cc|cc|c|}
\hline
\multirow{2}{*}{{Datasets}} & \multicolumn{2}{c|}{{Training Set}} & \multicolumn{2}{c|}{{Testing Set}} &
\multirow{2}{*}{{Ratio $\approx$}} \\
&  GAN-face  &  Real & GAN-face & Real & \\
\hline
 FFHQ-GAN-b & 2998 & 2991  & 750 & 748 & 1:1\\
\hline
 FFHQ-GAN-imb & 400 & 2000 & 100 & 500 & 1:5 \\
\hline
\end{tabular}
}
\label{tab:dataset}
\end{table}

\subsection{Implementation Details}
\label{sec:implement}

We implemented our method in PyTorch~\cite{paszke2019pytorch}. Experiments are conducted on a workstation with two NVIDIA GeForce 1080Ti GPUs. 

For iris detection, Mask R-CNN is trained using the datasets from \cite{wang2020iris,wang2021nir}. For each training eye image, the outer boundary mask of each iris is obtained using the method of \cite{wang2020iris} with default hyper-parameter settings. These masks are used to generate the iris bounding boxes and the corresponding masks for training, using the default settings in \cite{he2017mask}. 
In the test stage, given an input face image, we first use the face detector and landmark extractor of DLib~\cite{dlib09} to crop out the eye regions. 
Each cropped eye region is fed to Mask R-CNN for localizing the iris bounding box and segmentation mask. This process is repeated for both the left and right eyes to obtain the iris pairs as the input for our Residual Attention Network. We resize all iris pairs to a fixed size 96 $\times$ 96 for training and testing to ensure that the whole pipeline works well.

\begin{table}[t]
\caption{Results on the FFHQ-GAN dataset regarding its balanced (-b) and imbalanced (-imb) subsets.}
\centerline{
\resizebox{1.05\linewidth}{!}{
\setlength{\tabcolsep}{0.2em} 
\begin{tabular}{|c|c|ccccc|}
\hline
\multirow{2}{*}{Datasets} & \multirow{2}{*}{methods} & \multicolumn{5}{c|}{Metric}                                                                                                                                       \\ \cline{3-7} 
                          &                          & \multicolumn{1}{c|}{ACC}           & \multicolumn{1}{c|}{P}             & \multicolumn{1}{c|}{R}             & \multicolumn{1}{c|}{F1}            & AUC           \\ \hline
\multirow{4}{*}{FFHQ-GAN-b}   & ResNet                   & \multicolumn{1}{l|}{0.85}          & \multicolumn{1}{l|}{0.85}          & \multicolumn{1}{l|}{0.84}          & \multicolumn{1}{l|}{0.85}          & 0.92          \\ 
                          & Xception                 & \multicolumn{1}{l|}{0.95}          & \multicolumn{1}{l|}{0.95}          & \multicolumn{1}{l|}{0.96}          & \multicolumn{1}{l|}{0.95}          & 0.99          \\ 
                          & RAN with BCE loss       & \multicolumn{1}{l|}{0.92}          & \multicolumn{1}{l|}{0.92}          & \multicolumn{1}{l|}{0.92}          & \multicolumn{1}{l|}{0.92}          & 0.98          \\ \cline{2-7} 
                          & RAN with BCE+AUC loss   & \multicolumn{1}{l|}{\textbf{0.97}} & \multicolumn{1}{l|}{\textbf{0.98}} & \multicolumn{1}{l|}{\textbf{0.97}} & \multicolumn{1}{l|}{\textbf{0.97}} & \textbf{1.00} \\ \hline
\multirow{4}{*}{FFHQ-GAN-imb} & ResNet                   & \multicolumn{1}{l|}{0.72}          & \multicolumn{1}{l|}{0.28}           & \multicolumn{1}{l|}{0.41}          & \multicolumn{1}{l|}{0.33}          & 0.64          \\ 
                          & Xception                 & \multicolumn{1}{l|}{0.87}          & \multicolumn{1}{l|}{0.75}          & \multicolumn{1}{l|}{0.36}          & \multicolumn{1}{l|}{0.49}          & 0.85          \\  
                          & RAN with BCE loss       & \multicolumn{1}{l|}{0.89}          & \multicolumn{1}{l|}{0.84}           & \multicolumn{1}{l|}{0.41}          & \multicolumn{1}{l|}{0.55}          & 0.89          \\ \cline{2-7} 
                          & RAN with BCE+AUC loss   & \multicolumn{1}{l|}{\textbf{0.96}} & \multicolumn{1}{l|}{\textbf{0.89}} & \multicolumn{1}{l|}{\textbf{0.85}} & \multicolumn{1}{l|}{\textbf{0.87}} & \textbf{0.98} \\ \hline
\end{tabular}
}
}
\label{TAB_res_DT_b}
\vspace{-1em}
\end{table}

Table~\ref{Table:architectures} describes the details of our Residual Attention Network (RAN), where the Attention Module (AM) detailed in Figure~\ref{fig:attnet} is repeatedly stacked three times. 
The network is trained using Adam optimizer~\cite{kingma2015adam} with learning rate 0.001 and batch size 128. Training is terminated at 100 epochs for balanced data and 2,000 for imbalanced data. 

{\bf Hyper-parameters.} We set $p=2$ in Eq.~\eqref{eq:gamma} and $\gamma=0.4$ for balanced dataset, and $\gamma=0.6$ for imbalanced data. For the experiments on the balanced dataset, $\alpha$ in Eq.~\eqref{eq:jointloss} is set to 0.2. For the experiments on the imbalanced dataset, $\alpha$ is set to 0.4. These hyperparameters yields the best performance.

\subsection{Evaluation on the FFHQ-GAN dataset}
\label{sec:FFHQ:GAN:result}

We report evaluation of GAN-generated face detection on the FFHQ-GAN dataset in terms of Accuracy (ACC), Precision (P), Recall (R), F1 score (F1), the area under the curve (AUC) of the ROC, and Precision-Recall (PR) curves. 
Accuracy is calculated as $\text { ACC }=\frac{ \tp + \tn }{ \tp + \tn + \fp + \fn}$,
where $\fn$ and $\tn$ indicate false negatives and true negatives, respectively.
Precision-Recall is calculated as $\mbox{P}=\frac{ \tp }{ \tp + \fp }$ and $\mbox{R} =\frac{ \tp }{ \tp + \fn }$.
F1 score is the harmonic average value of P and R, as $\mbox{F1}=\frac{\mbox{2} \mbox{P} \mbox{R}}{\mbox{P} +\mbox{R} }$.

{\bf Results on the balanced and imbalanced set.}
To evaluate the effectiveness of the proposed method, we evaluate RAN trained with BCE + AUC loss against the ResNet-50~\cite{he2016deep} and Xception~\cite{chollet2017xception}, two of the widely-used DNN classification models trained with the BCE loss.
Table~\ref{TAB_res_DT_b} presents the classification results on both the balanced and imbalanced FFHQ-GAN datasets. 
The corresponding ROC and Precision-Recall curves are shown in Figure~\ref{fig_compareRes} (right), and the Confusion Matrices are shown in Figure~\ref{fig_cm_blance}.
Results show that the both the ResNet-50 and Xception obtain low Recall scores due to the imbalanced data distribution in Table~\ref{TAB_res_DT_b}.
In comparison, our method achieves the highest performance in all metrics. 
These results indicate that our method can effectively improve performance on both balanced and imbalanced data training. 
\subsection{Ablation Studies}
\label{sec:ablation}

{\bf Effect of the AUC loss.}
We compare the proposed RAN model trained with the ideal case with combined AUC and BCE loss in Eq.~\eqref{eq:jointloss} against the same model trained only with BCE loss.
Results are shown in Table~\ref{TAB_res_DT_b} and Figures~\ref{fig_compareRes} and \ref{fig_cm_blance}.
Observe that the proposed joint BCE+AUC loss outperforms the same model trained only with BCE loss alone in all evaluation metrics. In other words, the incorporation of the AUC loss improves the classification performance substantially and consistently.

\begin{figure}[t!]
\centerline{
  \includegraphics[width=0.8\linewidth]{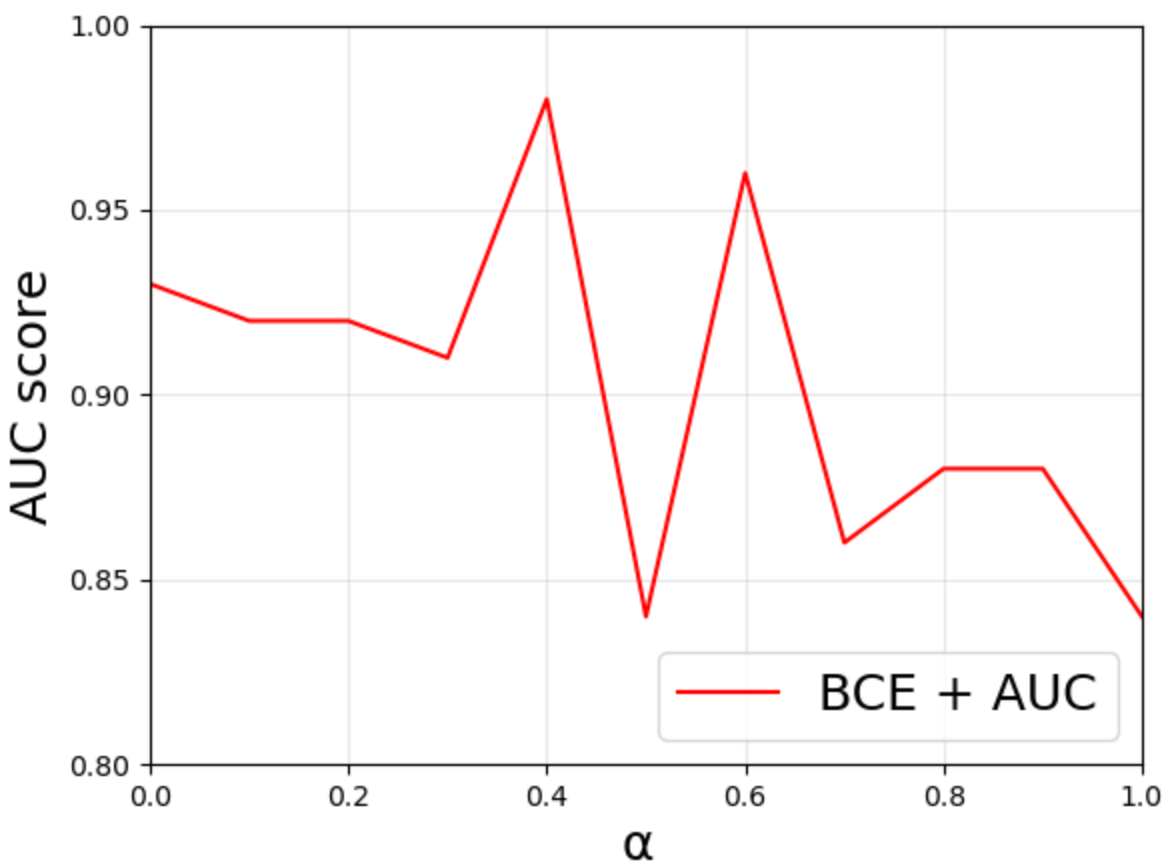}
    \vspace{-0.1cm}
}
\caption{{\bf Impact of hyperparameter $\alpha$} of the AUC loss  in Eq.\eqref{eq:gamma} for the GAN-generated face detection on the imbalanced dataset.
}
\label{fig:auchp}
\vspace{-0.5cm}
\end{figure}

{\bf Hyper-parameter analysis.}
We also study the impact of hyper-parameter $\alpha$ in our loss function in Eq.~\eqref{eq:jointloss} regarding detection performance of imbalanced data.
Figure~\ref{fig:auchp} shows the experimental results of the obtained AUC score versus $\alpha$ ranging from 0 to 1, and $\alpha=0.4$ yields the best detection performance. 

\begin{figure}[t]
\centerline{
{\footnotesize (a) Iris pairs from {\bf GAN-generated} faces}
}
\centerline{
  \includegraphics[width=\linewidth]{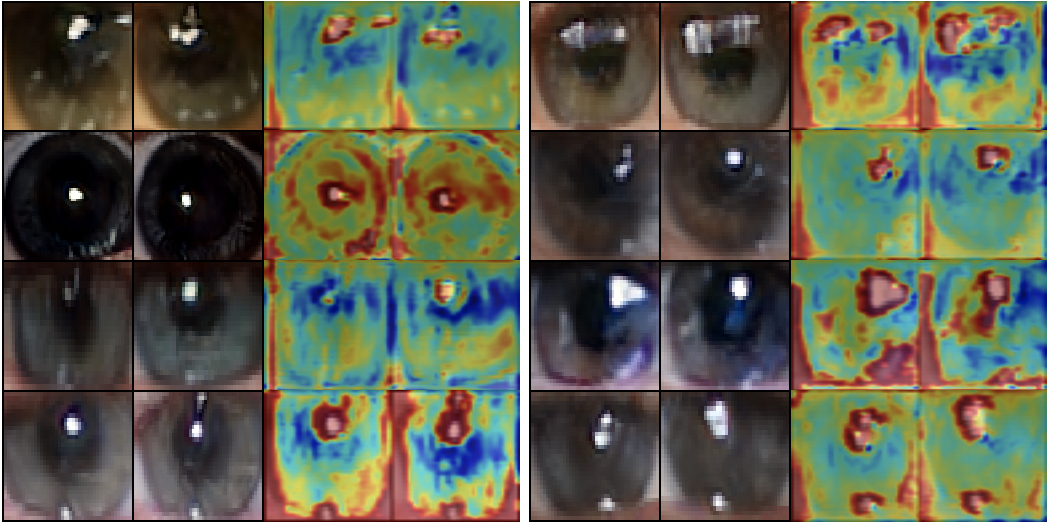}
}
\centerline{
{\footnotesize (b) Iris pairs from {\bf real} faces}}
\centerline{
  \includegraphics[width=\linewidth]{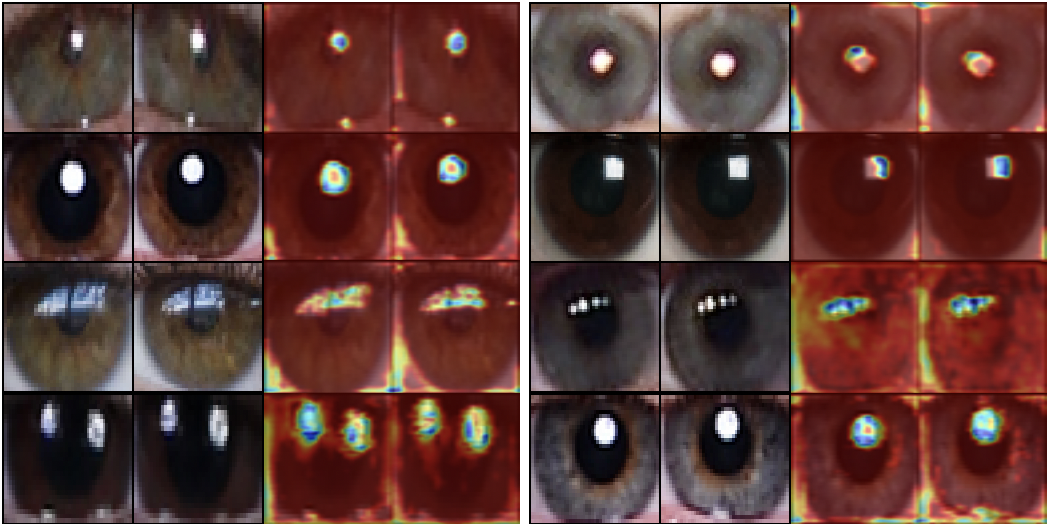}
}
\caption{Visualization of the {\bf extracted iris pairs} and the corresponding {\bf attention maps} obtained from our Residual Attention Network (RAN). Observe that the attention maps for GAN-generated faces better focus on the artifacts such as the corneal specular highlights, while the attention maps for real faces are widely distributed. This shows the effective learning of RAN for identifying GAN-generated faces. 
}
\label{attentionmaps}
\vspace{-4mm}
\end{figure}

\begin{figure*}[t]
\centerline{
\includegraphics[width=\linewidth]{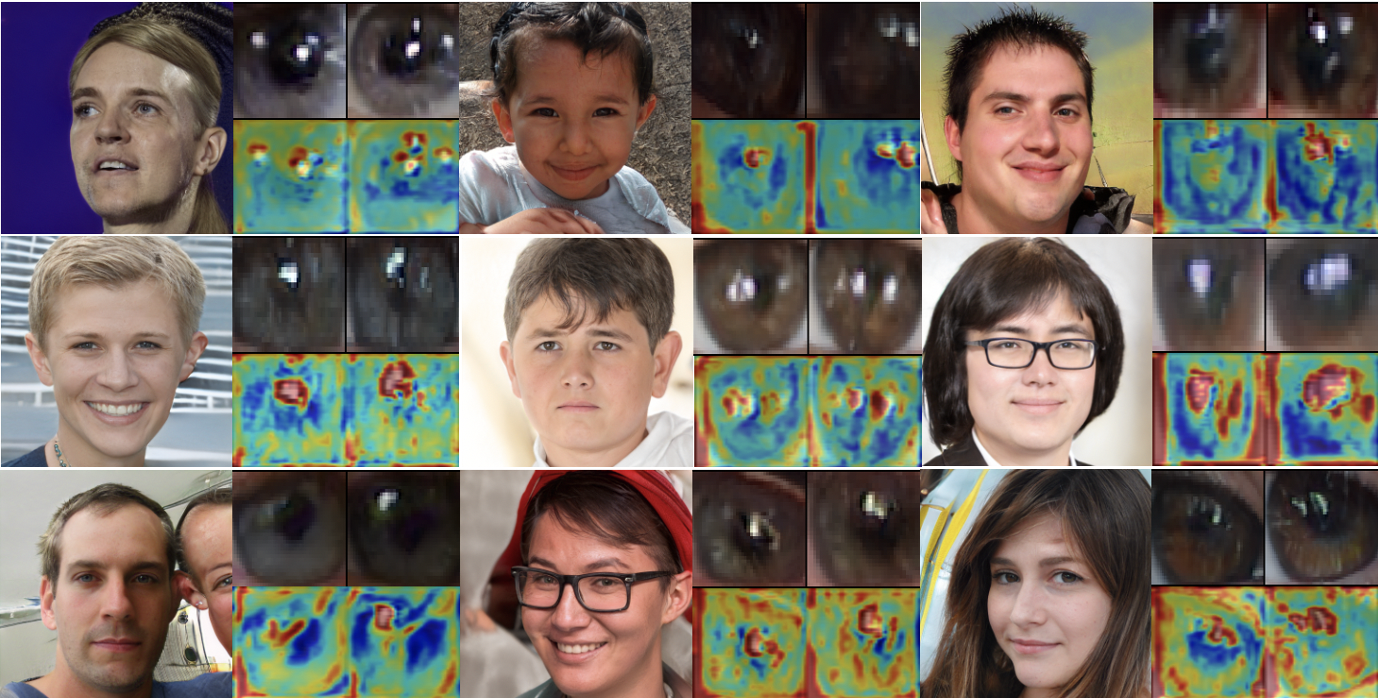}
}
\caption{Examples of {\bf detected GAN-generated faces} and their corresponding iris regions and the attention maps produced from our method. These examples show that our method can detect a wide range of face images, including those with tilted or side views where both irises are visible.
}
\label{faceandmaps}
\end{figure*}

\subsection{Qualitative Results}
\label{sec:qualitative}

Figure~\ref{attentionmaps} provides visualization of the attention maps of the real and GAN-generated iris examples. 
Observe that there is an obvious difference between the corresponding attention maps of the GAN-generated irises and the real ones. Concretely, the network attends on the whole iris part for the real images and attends to the highlight parts for the fake images. 
Figure~\ref{faceandmaps} shows additional examples of the GAN-generated face with the extracted iris pairs and corresponding attention maps. The visualization also provides an intuitive approach for human beings to identify GAN-generated faces by comparing their iris regions.

\section{Conclusion}
\label{sec:con}

In this work, we investigate the building a robust end-to-end deep learning framework for detecting GAN-generated faces. We show that GAN-generated faces can be distinguished from real faces by examining the consistency between the two iris regions. In particular, artifacts such as the corneal specular highlight inconsistencies can be robustly identified through end-to-end learning via the proposed Residual Attention Network. Our design of a joint loss combining the AUC loss with the cross-entropy loss can effectively deal with the learning from imbalanced data. We also showed that a direct optimization of the ROC-AUC loss is computationally not feasible, however relaxing the ROC AUC via the Wilcoxon-Mann-Whitney (WMW) statistics can provide a good approximation. 
Our GAN- face detection result is explainable, and the approach of spotting iris inconsistency can also serve as an useful cue for human users. Experimental results show that our model achieves superior performance on both balanced and imbalanced datasets for GAN-generated faces detection.




%




\bibliographystyle{IEEEtran}
\bibliography{main}

\vspace{-1cm}
\begin{IEEEbiography}
[{\includegraphics[width=1in,height=1.25in]{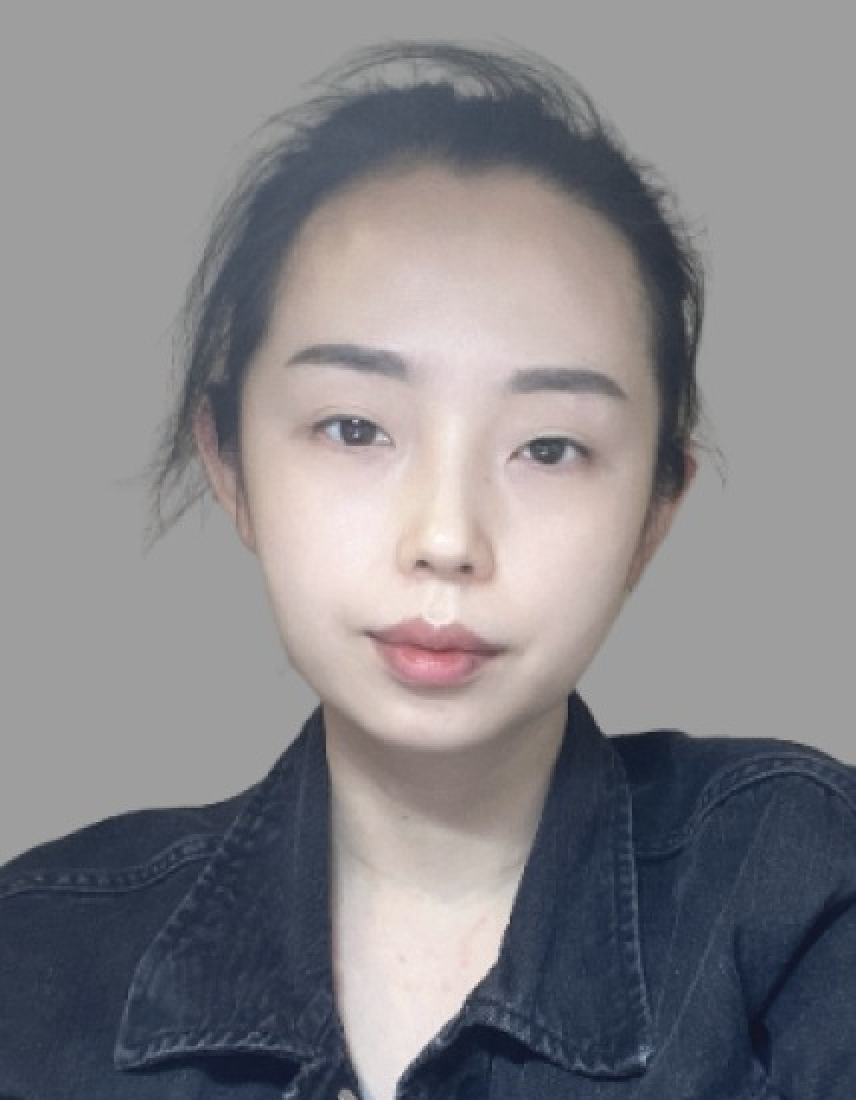}}]
{Hui Guo} 
is currently a Ph.D. candidate at University at Albany, the State University of New York. Her research interests include digital media forensics, computer vision, and deep learning.  
\vspace{-1cm}
\end{IEEEbiography}

\begin{IEEEbiography}
[{\includegraphics[width=1in,height=1.25in]{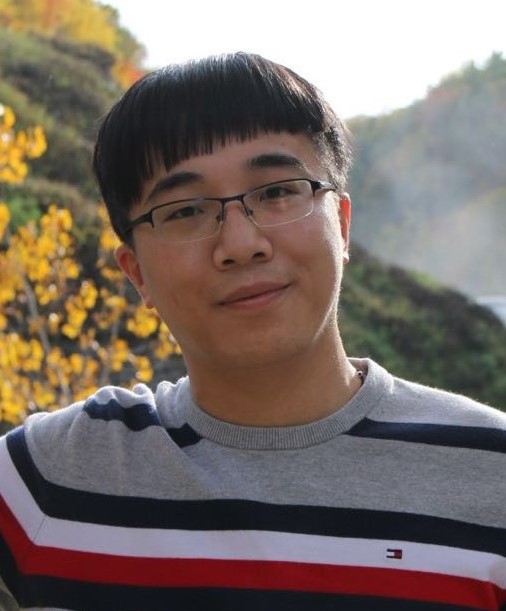}}]
{Shu Hu} is currently a fourth-year computer science Ph.D. candidate at University at Buffalo, the State University of New York. He received his M.A. degree in Mathematics from University at Albany, the State University of New York in 2020, and M.Eng. Degree in Software Engineering from University of Science and Technology of China in 2016. His research interests include machine learning, digital media forensics, and computer vision.  
\vspace{-1cm}
\end{IEEEbiography}

\begin{IEEEbiography}
[{\includegraphics[width=1in,height=1.25in]{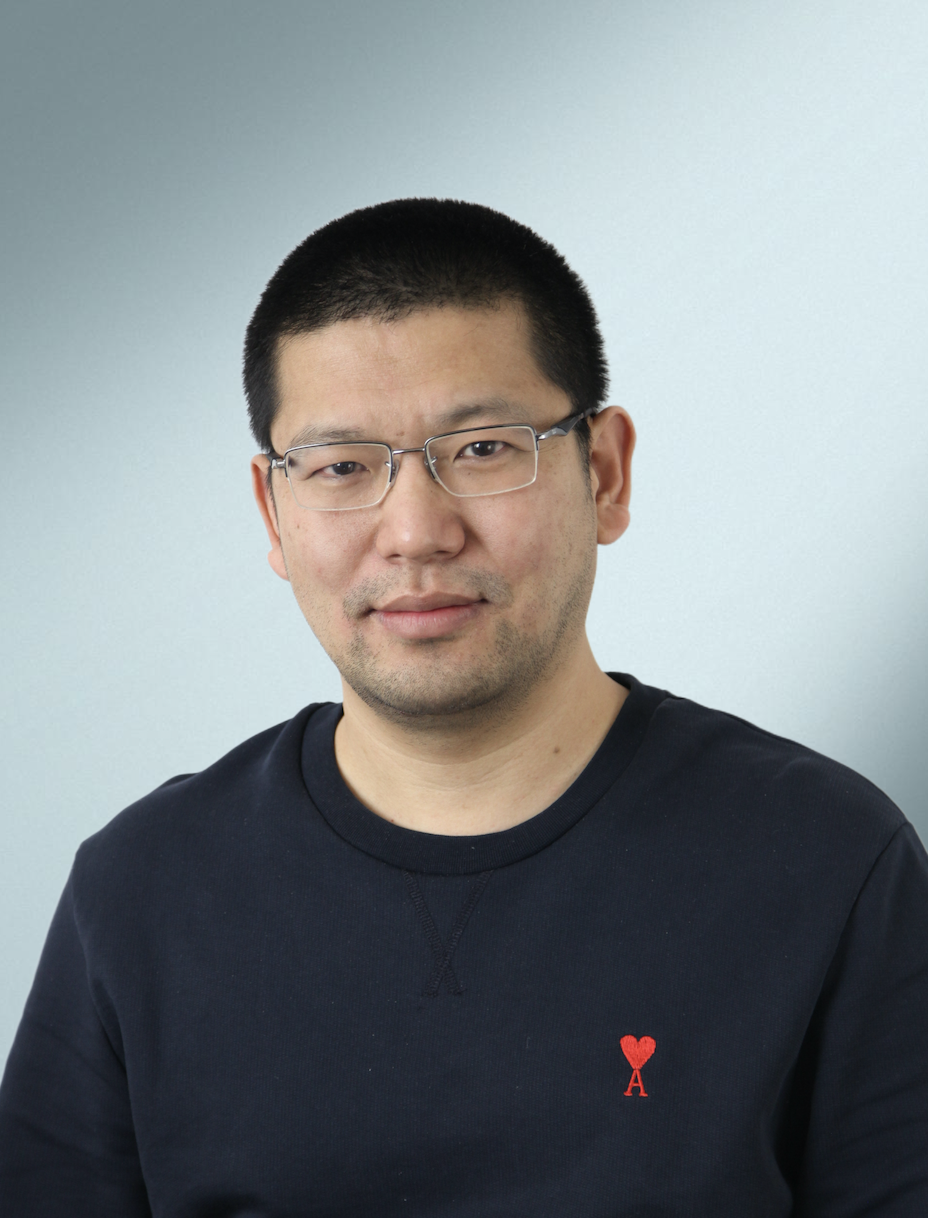}}]
{Xin Wang} (SM'2020) is currently a Senior Machine Learning Scientist at Keya Medical, Seattle, USA. He received his Ph.D. degree in Computer Science from the University at Albany, the State University of New York, NY in 2015. His research interests are in artificial intelligence, machine learning, reinforcement learning, medical image computing, computer vision, and media forensics. He is a senior member of IEEE. 
\vspace{-1cm}
\end{IEEEbiography}

\begin{IEEEbiography}
[{\includegraphics[width=1in,height=1.25in]{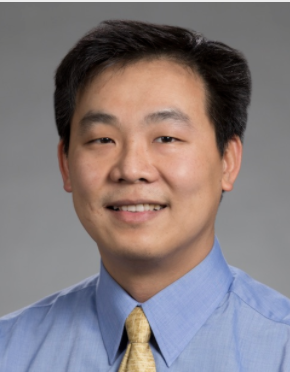}}]
{Ming-Ching Chang} is an Assistant Professor at the Department of Computer Science, College of Engineering and Applied Sciences (CEAS), University at Albany, State University of New York (SUNY). He was with the Department of Electrical and Computer Engineering from 2016 to 2018. During 2008-2016, he was a Computer Scientist at GE Global Research Center. He received his Ph.D. degree in the Laboratory for Engineering Man/Machine Systems (LEMS), School of Engineering, Brown University in 2008. He was an Assistant Researcher at the Mechanical Industry Research Labs, Industrial Technology Research Institute (ITRI) at Taiwan from 1996 to 1998. He received his M.S. degree in Computer Science and Information Engineering (CSIE) in 1998 and B.S. degree in Civil Engineering in 1996, both from National Taiwan University.
Dr. Chang’s expertise includes video analytics, computer vision, image processing, and artificial intelligence. His research projects are funded by GE Global Research, IARPA, DARPA, NIJ, VA, and UAlbany. He is the recipient of the IEEE Advanced Video and Signal-based Surveillance (AVSS) 2011 Best Paper Award - Runner-Up, the IEEE Workshop on the Applications of Computer Vision (WACV) 2012 Best Student Paper Award, the GE Belief - Stay Lean and Go Fast Management Award in 2015, and the IEEE Smart World NVIDIA AI City Challenge 2017 Honorary Mention Award. Dr. Chang serves as Co-Chair of the annual AI City Challenge CVPR 2018-2021 Workshop, Co-Chair of the IEEE Lower Power Computer Vision (LPCV) Annual Contest and Workshop 2019-2021, Program Chair of the IEEE Advanced Video and Signal-based Surveillance (AVSS) 2019, Co-Chair of the IWT4S 2017-2019, Area Chair of IEEE ICIP (2017, 2019-2021) and ICME (2021), TPC Chair for the IEEE MIPR 2022. He has authored more than 100 peer-reviewed journal and conference publications, 7 US patents and 15 disclosures. He is a senior member of IEEE and member of ACM.
\vspace{-1.5cm}
\end{IEEEbiography}

\begin{IEEEbiography}
[{\includegraphics[width=1in,height=1.25in]{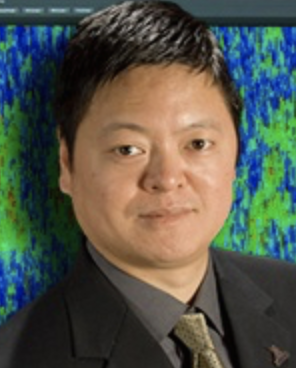}}]
{Siwei Lyu} is an SUNY Empire Innovation Professor at the Department of Computer Science and Engineering, the Director of UB Media Forensic Lab (UB MDFL), and the founding Co-Director of Center for Information Integrity (CII) of University at Buffalo, State University of New York. Dr. Lyu received his Ph.D. degree in Computer Science from Dartmouth College in 2005, and his M.S. degree in Computer Science in 2000 and B.S. degree in Information Science in 1997, both from Peking University, China.
Dr. Lyu’s research interests include digital media forensics, computer vision, and machine learning. Dr. Lyu is a Fellow of IEEE.
\end{IEEEbiography}

\EOD

 \end{document}